\pdfoutput=1

\documentclass[11pt]{article}

\usepackage{acl}

\usepackage{times}
\usepackage{latexsym}

\usepackage[T1]{fontenc}

\usepackage[utf8]{inputenc}

\usepackage{microtype}

\usepackage{bm}
\usepackage{amsmath}
\usepackage{amsfonts}
\usepackage{amssymb}
\usepackage{graphicx}
\usepackage{booktabs}
\usepackage{multirow}
\usepackage{makecell}
\usepackage{enumitem}
\usepackage{algorithm}
\usepackage{algorithmic}
\usepackage{xspace}
\usepackage{url}
\usepackage{utfsym}

\newcommand{\Action}[1]{\texttt{#1}}
\newcommand{\modelname}[0]{{\textsc{Fame}}\xspace}

\newcommand{\rw}[1]{\textcolor[rgb]{0.00,0.00,0.00}{#1}}

\title{Faithful Question Answering with Monte-Carlo Planning}

\author{Ruixin Hong\textsuperscript{1}, Hongming Zhang\textsuperscript{2}, Hong Zhao\textsuperscript{1}, Dong Yu\textsuperscript{2}, Changshui Zhang\textsuperscript{1} \\
\textsuperscript{1}Institute for Artificial Intelligence, Tsinghua University (THUAI); \\
\textsuperscript{1}Beijing National Research Center for Information Science and Technology (BNRist); \\
\textsuperscript{1}Department of Automation, Tsinghua University, Beijing, P.R.China \\
\textsuperscript{2}Tencent AI Lab, Seattle \\
\texttt{hrx20@mails.tsinghua.edu.cn,} 
\texttt{hongmzhang@tencent.com,} \\
\texttt{dyu@global.tencent.com,}
\texttt{zcs@mail.tsinghua.edu.cn,}
}

\begin{document}
\maketitle
\begin{abstract}
Although large language models demonstrate remarkable question-answering performances, revealing the intermediate reasoning steps that the models \textit{faithfully} follow remains challenging. 
In this paper, we propose \modelname (\underline{FA}ithful question answering with \underline{M}ont\underline{E}-carlo planning) to answer questions based on faithful reasoning steps.
The reasoning steps are organized as a structured entailment tree, which shows how premises are used to produce intermediate conclusions that can prove the correctness of the answer.
We formulate the task as a discrete decision-making problem and solve it through the interaction of a reasoning environment and a controller.
The environment is modular and contains several basic task-oriented modules, while the controller proposes actions to assemble the modules.
Since the search space could be large, we introduce a Monte-Carlo planning algorithm to do a look-ahead search and select actions that will eventually lead to high-quality steps. 
\modelname achieves \rw{advanced} performance on the standard benchmark.
It can produce valid and faithful reasoning steps compared with large language models with a much smaller model size.
\end{abstract}

\section{Introduction}

{
\setlength{\abovecaptionskip}{1mm}
\begin{figure}[t!]
\centering
\includegraphics[width=\columnwidth]{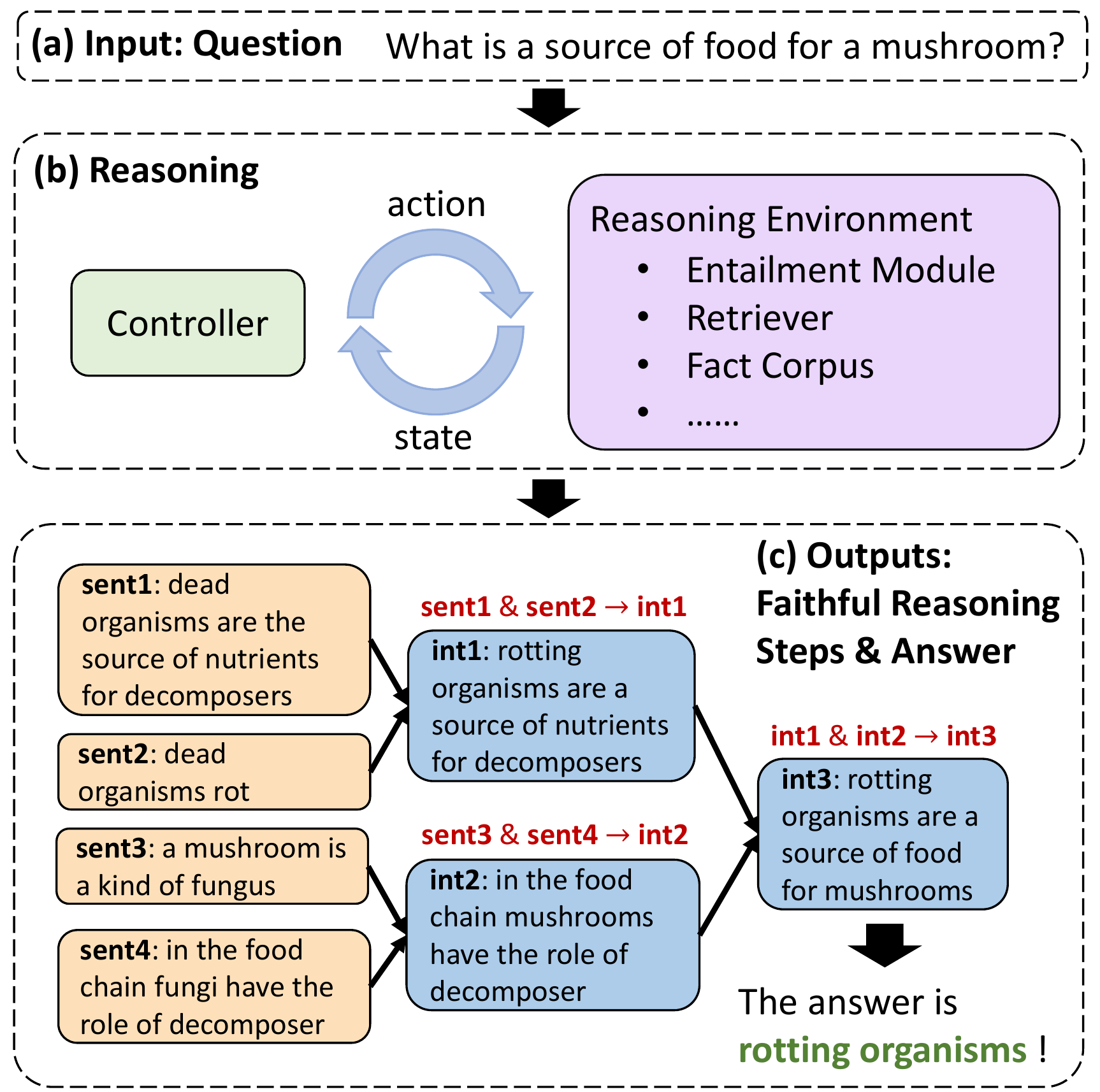}
\caption{Given a question, \modelname performs reasoning through the iterative interaction of a controller with a reasoning environment.
It produces the reasoning steps (in the form of an {entailment tree}) and the answer faithfully following from the steps.
The entailment tree contains the basic fact ($\mathrm{sent}_*$) and novel intermediate conclusions ($\mathrm{int}_*$) connected by entailment steps.
}
\vspace{-3mm}
\label{fig:intro}
\end{figure}
}
Enabling machines to reason and answer questions is a long-term pursuit in the AI community~\cite{mccarthy1960programs}.
In the field of question-answering (QA), large language models (LLMs) have achieved strong performances~\cite{DBLP:journals/corr/abs-2108-07258,DBLP:conf/nips/BrownMRSKDNSSAA20,DBLP:journals/corr/abs-2205-11916}.
However, the intermediate reasoning steps from the known premises to the answer are often implicit and invisible.
While some approaches encourage LLMs to produce the reasoning steps explicitly before generating the answer~\cite{DBLP:journals/corr/abs-2201-11903},
the answer may not \textit{faithfully} follow the intermediate steps,
i.e., the model could generate irrelevant or invalid steps while still resulting in the correct answer~\cite{DBLP:journals/corr/abs-2201-11903,DBLP:journals/corr/abs-2203-14465,DBLP:journals/corr/abs-2205-09712}.
Such a lack of faithfulness makes it difficult for users to trust the answers and debug the wrong answers, diminishing the overall trustworthiness of the QA systems.

To tackle this issue, the recently proposed faithful question-answering (FQA) task~\cite{entailer2022, DBLP:journals/corr/abs-2208-14271} asks the system to provide the reasoning steps that the answer faithfully follows, as demonstrated in Figure~\ref{fig:intro}(a) and~(c).
The reasoning steps are organized as an \textit{entailment tree}~\cite{DBLP:conf/emnlp/DalviJTXSPC21}, where each non-leaf node indicates an intermediate entailment step.
The provision of the faithful steps allows users to inspect and debug the system's reasoning process, potentially enabling the construction of interactive and teachable QA systems~\cite{DBLP:journals/corr/abs-2204-13074}.

Existing FQA methods typically adopt a step-wise approach to generate an entailment step at a time~\cite{entailer2022, DBLP:journals/corr/abs-2208-14271}.
When determining which step to generate next, they produce several candidate steps and select one based on the validity of the step.
However, they do not explicitly consider whether the selected step will ultimately result in a high-quality tree that supports the answer.
For complex questions that require multiple reasoning steps, such a lack of foresight might lead to the irreversible miss of the optimal step in a huge search space.
Furthermore, existing methods are based on either the model's internal beliefs or a small number of given facts, which limits their ability to ground to the external world and update their known facts.

In this paper, we propose \modelname, a novel FQA method integrating Monte-Carlo planning.
We formulate the task as a discrete decision-making problem and solve it through the interaction of a reasoning environment and a controller, as shown in Figure~\ref{fig:intro}.
The reasoning environment is modular.
We decompose the reasoning into several basic task-oriented modules, such as a retriever for updating known facts and a single-step entailment module for combining several premises to obtain a novel intermediate conclusion.
To assemble these modules, we leverage a controller 
(implemented with a generative language model) 
to observe the state of the environment and propose the next actions.
The final answer is derived based on the validity and faithfulness of the generated entailment tree.

To select actions foresightedly, we introduce a Monte-Carlo planning algorithm~\cite{DBLP:conf/ecml/KocsisS06} to do the look-ahead search.
Specifically, we assign an action value to each candidate action, which is iteratively updated based on the quality of its successor states after the action is explicitly executed.
With these values, we could make more informed decisions and select the actions that will eventually lead to a high-quality tree.
In addition, we design a verifier-guided iterative training technique to train the controller. 

Experiments on the standard benchmark EntailmentBankQA~\cite{DBLP:journals/corr/abs-2208-14271} show that \modelname outperforms previous best FQA methods by a large margin.
Manual evaluation results demonstrate that \modelname could produce valid and faithful reasoning steps compared with LLMs (i.e., GPT-3 and ChatGPT).
Further ablation results illustrate the advantages of Monte-Carlo planning compared to other planning methods.

{
\begin{table*}[th!]
\centering
\small
\resizebox{\textwidth}{!}{
\begin{tabular}{@{}lp{11cm}l@{}}
\toprule
Action & Effect & Example \\ \midrule
\Action{Retrieve}: <query>  
& Call a retriever to retrieve facts by <query> from the fact corpus $C$ and update candidate premises $X$. The query can be selected from $\{ H \} \cup X$ adaptively.
& \Action{Retrieve}: $\mathrm{int}_1$     \\ \midrule
\Action{Entail}: <premises>     
& Call an entailment module to generate a conclusion given the selected <premises> as input. Add a new step formed by the <premises> and the generated conclusion to ${T}_p$.
& \Action{Entail}:  $\mathrm{sent}_1$ \& $\mathrm{sent}_2$  \\ \midrule
\Action{End}: <is\_proved>      
& End reasoning and return whether the controller considers $H$ is proved. <is\_proved>  $\in$ \{``proved'', ``unproved''\}.
& \Action{End}: proved    \\ \bottomrule
\end{tabular}
}
\caption{
Action space for the controller. 
If the controller generates other text, it is treated as an invalid action.
}
\label{tab:action_space}
\end{table*}
}

\section{Related Work}

\textbf{Explicit Reasoning with Language Models.}
LLMs can achieve strong QA performances, even in few-shot and zero-shot settings~\cite{DBLP:journals/corr/abs-2108-07258,DBLP:conf/nips/BrownMRSKDNSSAA20,DBLP:journals/corr/abs-2205-11916}.
Recent approaches that explicitly generate the reasoning steps to derive the final answer~\cite{DBLP:journals/corr/abs-2203-14465,DBLP:journals/corr/abs-2203-11171,DBLP:journals/corr/abs-2205-10625,DBLP:journals/corr/abs-2204-02329} have shown significant improvement for many tasks~\cite{DBLP:journals/corr/abs-2210-09261}.
For example, Chain-of-Thought~\cite{DBLP:journals/corr/abs-2201-11903} encourages LLMs to generate several steps via few-shot prompting.
However, the generated steps could be unfaithful~\cite{DBLP:journals/corr/abs-2201-11903,DBLP:journals/corr/abs-2203-14465,DBLP:journals/corr/abs-2205-09712}.
To tackle this issue, faithful question-answering (FQA) proposes to answer questions based on the faithful reasoning steps~\cite{entailer2022, DBLP:journals/corr/abs-2208-14271,DBLP:journals/corr/abs-2209-07662,DBLP:journals/corr/abs-2201-06028}, where each step is a valid entailment and the intermediate conclusions support the answer.
Existing FQA methods typically adopt a step-wise approach.
To decide the next step, their strategy (e.g., overgenerate-and-filter of Entailer~\cite{entailer2022} and beam search of Selection-Inference (SI)~\cite{DBLP:journals/corr/abs-2208-14271}) might lack foresight and could be inadequate for complex questions.
Our work explicitly does the look-ahead search to make more foresighted decisions.
In addition, Entailer generates facts using the model's internal representation, which may hallucinate possibly incorrect facts to support its answer~\cite{entailer2022}.
SI requires a complete set of supporting facts for the question and can not update facts.
Our method is based on a deterministic corpus to prevent hallucination and could adaptively update facts using a retriever.

Our modular design is related to the approaches that break down the questions into smaller modules~\cite{DBLP:journals/corr/abs-2210-02406,DBLP:journals/corr/abs-2211-09066,DBLP:conf/acl/SanyalS022,DBLP:journals/corr/abs-2212-13894}.
In this paper, we compare against SI~\cite{DBLP:journals/corr/abs-2208-14271}, which is the representative modular FQA approach.

\textbf{Explanation for Question Answering.}
Great efforts have been devoted to improving the explainability of QA systems~\cite{DBLP:conf/nips/WiegreffeM21,DBLP:journals/corr/abs-2010-00389,DBLP:journals/tacl/LammPAACSC21,DBLP:journals/corr/abs-2112-07772,DBLP:journals/corr/abs-2210-12487}. 
Recently, EntailmentBank~\cite{DBLP:conf/emnlp/DalviJTXSPC21} proposes to formulate the reasoning steps of QA systems as multi-step textual entailments, which provides the most detailed and informative explanation.
A series of methods are developed to reconstruct such a tree-structured explanation for the correct answer~\cite{DBLP:conf/emnlp/DalviJTXSPC21,DBLP:conf/naacl/HongZYZ22,yang2022nlproofs,DBLP:conf/naacl/Ribeiro0MDWZCXH22,DBLP:journals/corr/abs-2210-17095}.
Our work is based on EntailmentBank but focuses on QA instead of post-hoc explanations.

\textbf{Monte-Carlo Planning for Language.}
Despite remarkable successes in games~\cite{DBLP:journals/nature/SilverSSAHGHBLB17,DBLP:journals/nature/SchrittwieserAH20}, few works attempt to apply Monte-Carlo planning (MCP) to language.
Previous works use MCP for decoding during language generation~\cite{DBLP:conf/ccnlg/KumagaiKMANN16,DBLP:journals/jaciii/KumagaiKMANN18,DBLP:conf/emnlp/LeblondASPLASV21,DBLP:conf/naacl/ChaffinCK22}, but how to use MCP for QA is still underexplored.
Our work formulates QA as a decision-making problem and explores the utilization of MCP.

\section{Task Definition}
Faithful question-answering~\cite{entailer2022, DBLP:journals/corr/abs-2208-14271} requires to answer the question and provide the valid reasoning steps that the answer faithfully follows.
The inputs include a question $Q$ and candidate options $O=\{o_1, o_2, \dots, o_{|O|})\}$.\footnote{For open-ended questions, we follow~\citet{entailer2022} to collect candidate options using an external source (e.g., Macaw~\cite{DBLP:journals/corr/abs-2109-02593})}
The desired outputs are valid reasoning steps ${T}$ in the form of the entailment tree and an answer $o_a \in O$, which follows ${T}$.
The entailment tree ${T}$ consists of multi-premise entailment steps, whose leaf nodes ($\mathrm{sent}_*$) are facts selected from a fact corpus $C$, intermediate nodes ($\mathrm{int}_*$) are novel intermediate conclusions.
A tree is considered \textit{valid} if each non-leaf node is a valid entailment of its immediate children,
and is considered \textit{faithful} if its conclusion of the root node can support the option $o_a$.

\section{Our Approach: \modelname}
Given a question $Q$ and options $O$, following previous works~\cite{entailer2022, DBLP:journals/corr/abs-2209-07662}, we first convert them into declarative hypotheses $\{H_1,\dots,H_{|O|}\}$.\footnote{We follow~\citet{entailer2022} to use a generation model whose input is $q+o_i$ and output is $H_i$.
Specifically, we use a T5-large~\cite{DBLP:journals/jmlr/RaffelSRLNMZLL20} trained on EntailmentBank.}
We then try to generate an entailment tree for each hypothesis in a forward chaining manner and select the most plausible option based on the validity and faithfulness of trees.

We propose to formulate the task as a discrete decision-making problem. 
The reasoning is done through several interactions between a reasoning environment and
a controller.
In each interaction, the controller observes a state from the environment and predicts the next action.
Then the environment executes the action and updates its state.\footnote{Illustrations of the reasoning process are in Appendix~\ref{sec:Illustrations}.}
Since the search space could be large, we introduce a Monte-Carlo planning algorithm to select the optimal action.
We introduce details about the environment, controller, and Monte-Carlo planning in Sec~\ref{sec:environment},~\ref{sec:controller}, and~\ref{sec:MCP}, respectively.

\subsection{Reasoning Environment} 
\label{sec:environment}

\noindent \textbf{State.}
A reasoning state $s = \{H, {T}_p, X \}$ consists of three parts: a target hypothesis $H$, a partial tree ${T}_p$, and candidate premises $X$.
${T}_p$ contains the entailment steps so far.
$X$ is the set of sentences that can be selected as premises for the next entailment step.
Sentences in $X$ are either facts retrieved from the corpus ($\mathrm{sent}_*$) or conclusions generated by previous steps ($\mathrm{int}_*$).
The maximum size of X is restricted to 25 to fit the input length limitation of common language models~\cite{DBLP:conf/emnlp/DalviJTXSPC21}.

\noindent \textbf{Action.}
We consider three types of action $a \in \mathcal{A}(s)$ for a state $s$, as shown in Table~\ref{tab:action_space}.
The entailment module is a seq2seq generation model to perform single-step entailment reasoning.
Implementation details of the environment can be found in Sec~\ref{Sec:Implementation_detail}.

\subsection{Reasoning Controller}
\label{sec:controller}
The controller is a sequence generation model whose input is a linearized state and whose outputs are actions.
The input sequence is the concatenation of $H$, ${T}_p$, and $X$.
The linearized ${T}_p$ is a series of steps, where the step premises are connected with ``$\&$'' and the step conclusion comes after ``$\rightarrow$.''
For each state, the controller predicts multiple candidate actions and their likelihood scores.

\subsection{Monte-Carlo Planning}
\label{sec:MCP}
To select the optimal actions, solely using the controller's likelihood scores could be insufficient.
We propose to select actions based on the qualities of the successor states after the execution of actions.
Specifically, we first introduce a state verifier to estimate the scores of states (Sec.~\ref{sec:state_score}).
Each candidate action is assigned an action value $Q$, which is updated iteratively by the planning algorithm.
Our algorithm is based on the Monte-Carlo tree search with an upper confidence bound.
The upper confidence bound helps to balance between exploiting the currently best actions to obtain precise estimates and exploring sub-optimal actions to ensure that no good actions are missed.
Given the sufficient computational cost, this method can converge asymptotically to the optimal actions in the single agent problem~\cite{DBLP:conf/ecml/KocsisS06,DBLP:journals/amai/Rosin11,DBLP:journals/nature/SchrittwieserAH20}.

\subsubsection{State Verifier}
\label{sec:state_score}
{
\begin{figure}[t!]
\centering
\includegraphics[width=\columnwidth]{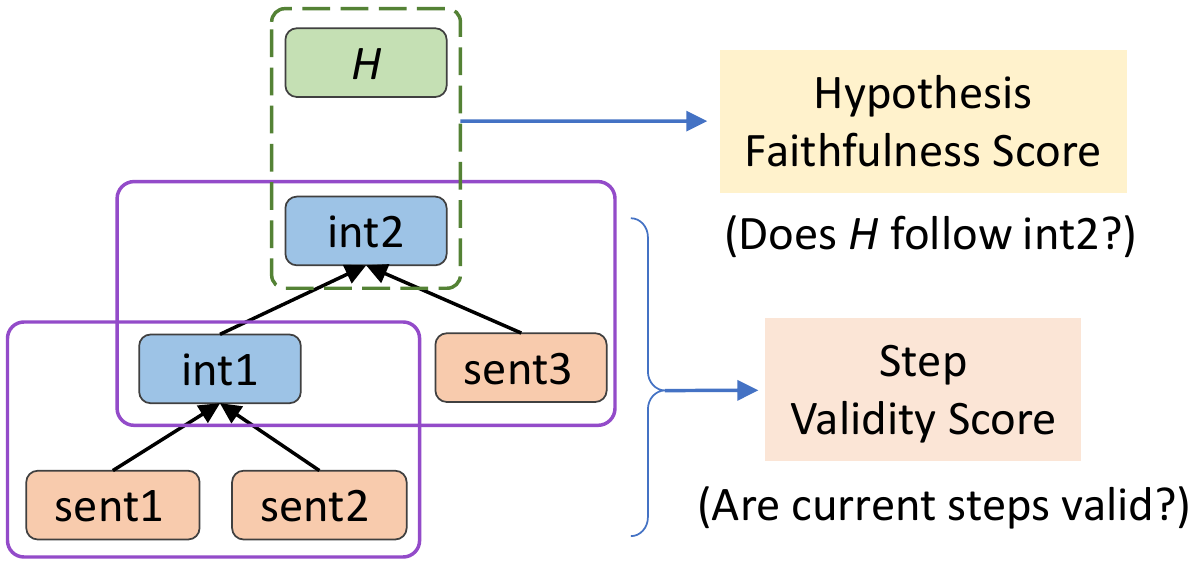}
\caption{
The step verifier $V$ scores a state based on its step validity and hypothesis faithfulness.
The hypothesis $H$ is the declarative form of the question and option.
}
\label{fig:state_score}
\end{figure}
}

A state is considered valuable if its ${T}_p$ is valid and ${T}_p$ can support the hypothesis $H$.
Thus, we introduce a state verifier $V$ to estimate the score of a state from two perspectives, as illustrated in Figure~\ref{fig:state_score}.
(1) \textbf{Step Validity}. Is ${T}_p$ valid?
We use a step verifier $V_{s}$ to check the validity of a single step.
The step verifier takes the premises and the conclusion as input and predicts a continuous score in $\left[ 0,1 \right]$ indicating how likely the conclusion is to be entailed by the premises.
Then the step scores are aggregated to produce the validity score of ${T}_p$, 
{
\setlength{\abovedisplayskip}{3pt}
\setlength{\belowdisplayskip}{3pt}
\setlength{\abovedisplayshortskip}{3pt}
\setlength{\belowdisplayshortskip}{3pt}
\begin{equation}
\label{equ:valid}
\small
    \texttt{Valid}({T}_p) = \frac{1}{|{T}_p|} \sum_{\text{step} \in {T}_p} V_{s}(\text{step}).
\end{equation}
}(2) \textbf{Hypothesis Faithfulness}. Does $H$ faithfully follow ${T}_p$?
To check this, we extract the highest conclusion $\mathrm{int}_{h}$ in ${T}_p$ (e.g., $\mathrm{int}_{2}$ in Figure~\ref{fig:state_score}) and verify if $\mathrm{int}_{h}$ can support $H$.
Following~\citet{DBLP:conf/emnlp/DalviJTXSPC21}, we use \texttt{BLEURT}~\cite{DBLP:conf/acl/SellamDP20} as the scorer $V_h$ to estimate the sentence similarity between $\mathrm{int}_{h}$ and $H$.
In addition, we check whether $H$ is entailed by $\mathrm{int}_{h}$ with the step verifier $V_s$.
{
\setlength{\abovedisplayskip}{3pt}
\setlength{\belowdisplayskip}{3pt}
\setlength{\abovedisplayshortskip}{3pt}
\setlength{\belowdisplayshortskip}{3pt}
\begin{equation}
\label{equ:faithful}
\small
    \texttt{Faithful}({T}_p, H) = (V_h(\mathrm{int}_{h}, H) + V_s(\mathrm{int}_{h} \rightarrow H))/2.
\end{equation}
}If more than one $\mathrm{int}_{h}$ exists, we take their maximum faithfulness score.
The overall state score is composed of its validity and faithfulness scores,
{
\setlength{\abovedisplayskip}{3pt}
\setlength{\belowdisplayskip}{3pt}
\setlength{\abovedisplayshortskip}{3pt}
\setlength{\belowdisplayshortskip}{3pt}
\begin{equation}
\label{equ:verifier}
\small
    V(s) = 
    (\texttt{Valid}({T}_p) + \texttt{Faithful}({T}_p, H))/2.
\end{equation}
}If there is no step yet (${T}_p=\emptyset$), then $V(s) = 0$.

\subsubsection{Planning Algorithm}
\label{sec:planning_algorithm}
{
\begin{figure*}[t!]
\centering
\includegraphics[width=\textwidth]{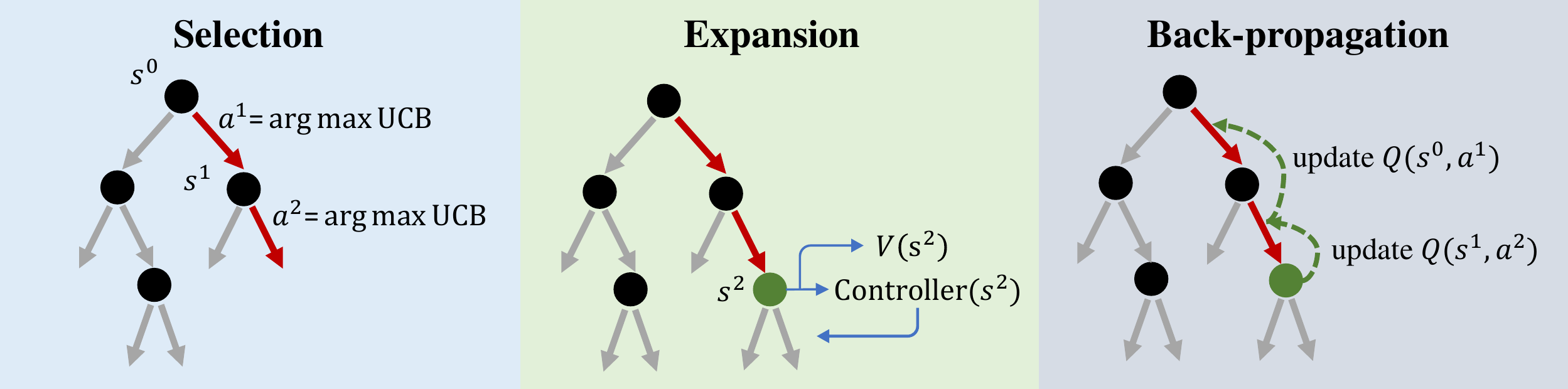}
\caption{The illustration of Monte-Carlo planning.
We construct a planning tree where each node is a state $s$.
\textbf{Selection}: In each simulation, we traverse the planning tree from $s^0$ by selecting actions with the maximum upper confidence bound (UCB, equation~(\ref{equ:ucb})).
\textbf{Expansion}: We expand a new node and evaluate the new state with the state verifier $V$.
The candidate actions of the new state are predicted by the controller and stored.
\textbf{Back-propagation}: Action values $Q$ are updated based on the scores of successor states after the action is executed.
For simplicity, here we assume that there are only two candidate actions for each state.
}
\label{fig:mcts}
\end{figure*}
}

Figure~\ref{fig:mcts} illustrates our planning algorithm.
We construct a planning tree where each node in the planning tree is correlated with a state $s$.
For each action $a$ of $s$, we store a set of statistics $\{ P(s,a), Q(s,a), N(s,a), \textit{Tra}(s,a) \}$,
where $P$ is the prior score (i.e., the likelihood score from the controller),
$Q$ is the action value,
$N$ is the visit count,
and $\textit{Tra}$ is the deterministic state transition.
$Q(s,a)$ can be viewed as the a posteriori score, which considers the future impact of $a$.
The planning algorithm ends after a fixed budget of simulations.
Each simulation consists of three stages.

\noindent $\bullet$ \textbf{Selection.}
The selection stage starts from the initial state $s^0$ (root node of the planning tree) and finishes when it reaches an unexpanded leaf node $s^m$.
For each $k=1,\dots,m+1$, an action $a^k$ is selected based on the statistics of $s^{k-1}$ to maximize an upper confidence bound~\cite{DBLP:journals/nature/SilverSSAHGHBLB17},
{
\setlength{\abovedisplayskip}{3pt}
\setlength{\belowdisplayskip}{3pt}
\setlength{\abovedisplayshortskip}{3pt}
\setlength{\belowdisplayshortskip}{3pt}
\begin{equation}
    \small
    a^k = \arg \max_{a \in \mathcal{A}(s)} \left[ Q(s,a)+ c_p P(s,a) \frac{\sqrt{\sum N(s,\cdot)}}{1+N(s,a)} \right] ,
    \label{equ:ucb}
\end{equation}
}where $\sum N(s,\cdot)$ is the total visit counts of $s$ and $c_p$ is a constant controlling the influence of the prior $P(s,a)$ over $Q(s,a)$.
With the upper confidence bound, the selection strategy initially favors actions with high prior scores and low visit counts (the second part) and gradually prefers actions with high action values (the first part).
The next states are looked up through the state transition $\textit{Tra}$.

\noindent $\bullet$ \textbf{Expansion.}
For the leaf node $s^m$ and the selected action $a^{m+1}$, we execute $a^{m+1}$ and get the next state $s^{m+1}$.
A new node correlated with $s^{m+1}$ is added to the planning tree.
We record the transition $\textit{Tra}(s^m, a^{m+1}) = s^{m+1}$ and the state score $V(s^{m+1})$.
We then use the controller to predict the candidate actions $a$ and their prior scores $p_a$.
Each action is initialized with $\{P(s,a)\!=\!p_a,\!Q(s,a)\!=\!0,\!N(s,a)\!=\!0\}$.
Note that during each simulation, we only execute one action and call the controller, environment, and state verifier at most once.

\noindent $\bullet$ \textbf{Back-propagation.}
The statistics along the trajectory $(s^0, a^1, \dots, s^{m}, a^{m+1})$ are updated.
For the final action, we set $Q(s^{m},a^{m+1})\!=\!V(s^{m+1})$.
For $k=m,\!\dots\!,1$, we update $Q(s^{k-1},a^{k})$ by integrating information from its next state $s^{k}$.
We consider $Q(s^{k-1},a^{k})$ as the probability that $a^{k}$ can correctly solve the task.
Given that $a^{k}$ can solve the task only requires one action of $s^{k}$ can solve the task,
the value of $a^{k}$ should be the disjunction of all actions of $s^{k}$.
We use the maximization operator to soften the disjunction\footnote{We follow the T-conorm of G\"odel fuzzy logic~\cite{DBLP:journals/sLogica/CaicedoR10, gupta1991theory}.} and obtain the value estimation $G$ in this simulation,

{
\setlength{\abovedisplayskip}{-5pt}
\setlength{\belowdisplayskip}{3pt}
\setlength{\abovedisplayshortskip}{-5pt}
\setlength{\belowdisplayshortskip}{3pt}
\begin{equation}
\small
    G(s^{k-1},a^{k}) = \bigvee_{a \in \mathcal{A}(s^k)} Q(s^k,a) = \max_{a \in \mathcal{A}(s^k)} Q(s^k,a).
\end{equation}
}We update the statistics as follows,
{
\setlength{\abovedisplayskip}{3pt}
\setlength{\belowdisplayskip}{-5pt}
\setlength{\abovedisplayshortskip}{3pt}
\setlength{\belowdisplayshortskip}{-5pt}
\begin{equation}
\small
\begin{aligned}
        Q(s^{k-1},a^{k}) & \leftarrow \frac{N(s^{k-1},a^{k}) \cdot Q(s^{k-1},a^{k}) + G(s^{k-1},a^{k})}{N(s^{k-1},a^{k}) + 1}, \\
    N(s^{k-1},a^{k}) & \leftarrow N(s^{k-1},a^{k})+1 .
\end{aligned}
\end{equation}
}

\subsubsection{Option Selection}
At the end of the planning, an extra selection stage is run to get the best state $s_{\text{best}}$ using equation~(\ref{equ:ucb}).
We score the option $o$ corresponding to $H$ by aggregating the verifier's and controller's judgments,
{
\setlength{\abovedisplayskip}{3pt}
\setlength{\belowdisplayskip}{3pt}
\setlength{\abovedisplayshortskip}{3pt}
\setlength{\belowdisplayshortskip}{3pt}
\begin{equation}
    \small
    \texttt{Score}(o) = (V(s_{\text{best}}) + P(s_{\text{best}}, {\text{\Action{End}: proved}}))/2,
\end{equation}
}where $P(s_{\text{best}}, {\text{\Action{End}: proved}})$ is the action likelihood score from the controller.
With all options scored, we eventually select the option with the highest score as the answer.

\subsection{Controller Training}
Given a correct trajectory $(s^0, a^1, \dots, s^{m}, a^{m+1})$, the controller is trained to maximize the likelihood of the correct actions $P(s^{k-1}, a^{k})$.
We first train the controller with behavior cloning~\cite{DBLP:journals/corr/abs-2112-09332} to imitate an oracle strategy.
Then, we iteratively use the controller to generate trajectories and fine-tune the controller with high-quality trajectories selected by the verifier.

\noindent \textbf{Behavior Cloning.}
Given a state, the oracle strategy gives an action based on the gold entailment tree:
(1) If the hypothesis $H$ is already in $X$, return \Action{End}: proved;
(2) Otherwise, if $X$ contains all premises of the next step, return \Action{Entail}: <premises>;
(3) Otherwise, select a query from $\{ H \} \cup X$ such that the updated $X$ contains as many leaf facts of the gold tree as possible, and return \Action{Retrieve}: <query>.

\noindent \textbf{Verifier-Guided Iterative Training.}
Training with behavior cloning alone may have two problems.
First, since the entailment trees are annotated only for the correct options, we do not have training trajectories for the wrong options.
Second, the trajectory for the correct option may not be unique.
To tackle these problems, we propose an iterative training process.
\textbf{Step 1:}
We train a controller using behavior cloning on the correct options.
\textbf{Step 2:}
We use the controller to generate trajectories for all training options.
For the correct option, 
we check the quality of the trajectory with the state verifier.
If the state score of the final state is greater than a threshold, we add the trajectory to the training data.
In this way, we could find some potentially correct trajectories, which may not be the same as the annotated ones but can still produce valid and faithful entailment trees.
For the wrong option,
we modify each $(s^{k-1}, a^{k})$ to $(s^{k-1}, \Action{End}\!:\!\text{unproved})$ and add them to the training data. 
\textbf{Step 3:}
We fine-tune the controller with the updated training data and go to Step 2 until the performance is saturated.

\section{Experiments}

\subsection{Dataset}
We conduct experiments based on EntailmentBank~\cite{DBLP:conf/emnlp/DalviJTXSPC21}.
It consists of 1,840 expert-annotated entailment trees, each of which corresponds to a hypothesis derived from a question+\textit{correct option} from the ARC~\cite{DBLP:journals/corr/abs-1803-05457} dataset.
The leaf facts of trees are selected from the WorldTreeV2 corpus~\cite{DBLP:conf/lrec/XieTMWMJ20}.

Since EntailmentBank is originally designed for post-hoc tree reconstruction instead of QA, we follow~\citet{DBLP:journals/corr/abs-2208-14271} to convert it to EntailmentBankQA by adding back the 4-way multiple options from the ARC dataset.
Summary statistics are shown in Table~\ref{tab:EntailmentBankQA_data}.
Answering these questions requires retrieving from a large fact corpus and complex multi-hop reasoning (each tree contains 7.6 nodes and 3.2 steps on average).

\subsection{Implementation Details}
\label{Sec:Implementation_detail}

\noindent \textbf{Retriever.}
The retriever is based on the Siamese Network encoder provided by Sentence Transformers~\cite{DBLP:conf/emnlp/ReimersG19}.
We fine-tune the {all-mpnet-base-v2} encoder via a contrastive loss~\cite{DBLP:journals/corr/abs-1807-03748} to maximize the cosine similarity between the hypothesis and its leaf facts of the tree from the EntailmentBank training split.
We use the fact corpus provided by EntailmentBank following previous works.

\noindent \textbf{Entailment Module.}
The entailment module is a T5-large model which takes the premises as input and generates the conclusion.
Following MetGen~\cite{DBLP:conf/naacl/HongZYZ22}, we divide the single-step entailment reasoning into a set of basic logical reasoning (i.e., substitution, conjunction, and if-then).
Special prefixes are used to specify the reasoning type the model should perform.
We train the module with entailment steps from the EntailmentBank training split.
Given premises, we generate conclusions using all types of modules and select the conclusion with the highest step verifier score.

\noindent \textbf{Step Verifier $V_s$.}
We fine-tune a DeBERTa-large model~\cite{DBLP:journals/corr/abs-2111-09543} to classify if a step is valid.
We use the human-labeled data provided by~\citet{entailer2022}, which contains 1,827 valid steps and 1,564 invalid steps.
We also include 4,175 valid steps from the EntailmentBank training split.

\noindent \textbf{Controller.}
The controller is implemented with a T5-large model.
For each state, we use the beam search to generate five candidate actions and then exclude the ill-formed and invalid actions (e.g., \Action{Retrieve:} $\mathrm{int}_2$, but $\mathrm{int}_2$ is not yet available in $X$).
For iterative training, we use a threshold of 0.98 to select high-quality trajectories.
Model performance is typically saturated after five iterations.

\noindent \textbf{Planning Algorithm.}
We select the hyperparameters with dev split (Appendix~\ref{sec:mcts_para}).
$c_p$ in equation~(\ref{equ:ucb}) is 0.2.
The simulation/action budget is 30.
Note that we execute only one action in each simulation.
We run our method three times and report the mean and standard deviation.
More implementation details can be found in Appendix~\ref{sec:appendix_implementation_details}.

{
\begin{table}[t]
\small
\centering
\begin{tabular}{@{}l|ccc|c@{}}
\toprule
 & Train & Dev & Test & All \\ \midrule
Questions & 1,313 & 187 & 340 & 1,840 \\
\quad Easy & 920 & 128 & 234 & 1,282 \\
\quad Chal (Challenge) & 393 & 59 & 106 & 558 \\
Entailment Steps & 4,175 & 597 & 1,109 & 5,881 \\ \bottomrule
\end{tabular}%
\caption{EntailmentBankQA Statistics.}
\label{tab:EntailmentBankQA_data}
\end{table}
}

\subsection{Baselines}
We compare with recent SOTA FQA methods.
\textbf{Entailer}~\cite{entailer2022} uses the model's internal beliefs to produce trees and answers.
For each hypothesis, it generates six candidate steps and selects the best one with a verifier.
\textbf{NELLIE}~\cite{DBLP:journals/corr/abs-2209-07662} is a backward-chaining inference engine based on a semi-structured fact base and extra labeled inference templates.
\textbf{Selection-Inference (SI)}~\cite{DBLP:journals/corr/abs-2208-14271} iteratively generates trees by a selection model and an inference model, and produces answers with a halter model.
For Entailer, we use its open-source T5-large model and default parameters.\footnote{\url{https://allenai.org/data/entailer}}
Since it also needs hypothesis conversion, we use the same conversion results as ours for a fair comparison.

\section{Result Analysis}

\subsection{Faithful Question Answering}

\noindent \textbf{EntailmentBankQA.}
As shown in Table~\ref{tab:EntailmentBankQA}, \modelname outperforms baseline methods by a large margin, improving the accuracy from 53.1\% to 67.1\%.
We also experiment in the settings of~\citet{DBLP:journals/corr/abs-2208-14271}, where a small set of facts is provided.
Task 1 provides exactly the leaf facts of the gold tree for the correct option, while Task 2 additionally includes several distractors.
Since retrieval is prohibited and not required, we remove the retrieval from the action space.
Table~\ref{tab:EntailmentBankQA_task12} shows the results.
\modelname could be adapted to these settings and achieve higher accuracy than SI, which is based on larger language models.
The errors in \modelname are traceable. 
The most frequent cause of errors is the mistakes in the intermediate steps (See the error analysis in Appendix~\ref{sec:error_analysis}).

{
\begin{table}[t]
\centering
\small
\begin{tabular}{@{}lccc@{}}
\toprule
Method & All & Easy & Chal \\ \midrule
NELLIE~\dag & 43.7 & 45.5 & 39.6 \\
Entailer & 53.1 & 56.4 & 46.2 \\ \midrule
\modelname (Ours) & \textbf{67.1}$\pm$0.6 & \textbf{70.8}$\pm$0.4 & \textbf{59.1}$\pm$1.2 \\ \bottomrule
\end{tabular}
\caption{
Answer accuracy (\%) on the EntailmentBankQA test split.
\dag~indicates results from the published paper.
All methods are based on T5-large.}
\label{tab:EntailmentBankQA}
\end{table}
}

{
\begin{table}[t]
\centering
\small
\begin{tabular}{@{}lcc@{}}
\toprule
Method & Task 1 & Task 2 \\ \midrule
SI+Halter~\dag & 72.4 & 55.9 \\
SI+Halter+Search~\dag & 83.2 & 72.9 \\ \midrule
\modelname (Ours) & \textbf{91.5}$\pm$0.8 & \textbf{78.2}$\pm$0.9 \\ \bottomrule
\end{tabular}
\caption{
Task 1 and Task 2 accuracy (\%) on the EntailmentBankQA test split.
\dag~indicates results from SI.
SI is based on Chinchilla-7B~\cite{DBLP:journals/corr/abs-2203-15556}.
}
\label{tab:EntailmentBankQA_task12}
\end{table}
}

{
\begin{table}[t]
\centering
\small
\resizebox{\columnwidth}{!}{
\begin{tabular}{@{}lcccc@{}}
\toprule
\multicolumn{1}{l}{\multirow{2}{*}{Method}} & \multicolumn{3}{c}{WorldTreeQA} & \multirow{2}{*}{OBQA} \\ \cmidrule(lr){2-4}
\multicolumn{1}{c}{} & All & Easy & Chal &  \\ \midrule
NELLIE~\dag & 38.3 & 40.8 & 32.3 & - \\
Entailer & 50.7 & 54.4 & 41.9 & 45.6 \\ \midrule
\modelname (Ours) & \textbf{61.5}$\pm$0.4 & \textbf{65.1}$\pm$0.4 & \textbf{52.6}$\pm$0.3 & \textbf{46.6}$\pm$0.4 \\ \bottomrule
\end{tabular}
}
\caption{Cross-dataset results on the WorldTreeQA and OBQA test split.}
\label{tab:cross-dataset}
\end{table}
}
 
\noindent \textbf{Cross-Dataset Performance.}
To evaluate the generalization capability of our method, we conduct experiments on WorldTreeQA~\cite{DBLP:conf/lrec/XieTMWMJ20} and OBQA~\cite{DBLP:conf/emnlp/MihaylovCKS18} following previous works~\cite{DBLP:journals/corr/abs-2209-07662,entailer2022}.
We evaluate on the test split of WorldTreeQA (1,177 easy and 487 challenge questions, no overlap with EntailmentBankQA) and OBQA (500 questions) without further fine-tuning on their training split.
We use the fact corpus that is included with the dataset.
As shown in Table~\ref{tab:cross-dataset}, \modelname achieves better cross-dataset performances.

\subsection{Reasoning Step Quality}

\noindent \textbf{Automatic Evaluation on EntailmentBank.}
To investigate whether our method can generate a valid entailment tree, we first perform the automatic evaluation on EntailmentBank, where we generate trees for correct options.
The validity of the generated tree is evaluated by comparing its leaves, step structures, and intermediate conclusions against the gold one.
The F1 score is computed, and the AllCorrect score is 1 if F1=1, otherwise 0.
The Overall AllCorrect metric measures whether the generated tree and the gold tree are identical.\footnote{As discussed by~\citet{yang2022nlproofs} and~\citet{DBLP:conf/naacl/HongZYZ22}, these automatic metrics do not account for the existence of multiple valid trees.
And the Overall AllCorrect score is a very harsh metric, but it is the fairest metric we could use.
}
Please refer to Appendix~\ref{sec:auto_tree_metric} for more evaluation details.
We use the controller trained with behavior cloning on the correct options.
We compare with the SOTA methods that are specifically designed for tree reconstruction.
As shown in Table~\ref{tab:tree_generation}, \modelname achieves competitive performances in all metrics, indicating that it could generate valid and high-quality trees.

{
\begin{table*}[t]
\centering
\small
\resizebox{\textwidth}{!}{
\begin{tabular}{@{}lccccccc@{}}
\toprule
\multicolumn{1}{c}{\multirow{2}{*}{Method}} & \multicolumn{2}{c}{Leaves} & \multicolumn{2}{c}{Step Structures} & \multicolumn{2}{c}{Intermediates} & \multirow{1}{*}{Overall} \\
\multicolumn{1}{c}{} & F1 & AllCorrect & F1 & AllCorrect & F1 & AllCorrect & AllCorrect  \\ \midrule
EntailmentWriter~\cite{DBLP:conf/emnlp/DalviJTXSPC21}~\dag & 35.7 & 2.9 & 6.1 & 2.4 & 33.4 & 7.7 & 2.4 \\
MetGen~\cite{DBLP:conf/naacl/HongZYZ22}~\dag & 34.8 & 8.7 & 9.8 & 8.6 & 36.6 & 20.4 & 8.6 \\
NLProofs~\cite{yang2022nlproofs}~\dag & 43.2 & 8.2 & 11.2 & 6.9 & \textbf{42.9} & 17.3 & 6.9 \\
RLET~\cite{DBLP:journals/corr/abs-2210-17095}~\dag & 38.3 & 9.1 & 11.5 & 7.1 & 34.2 & 12.1 & 6.9 \\
IRGR~\cite{DBLP:conf/naacl/Ribeiro0MDWZCXH22}~\dag & \textbf{45.6} & 11.8 & 16.1 & 11.4 & 38.8 & \textbf{20.9} & 11.5 \\ \midrule
\modelname (Ours) & 43.4$\pm$0.3 & \textbf{13.8}$\pm$0.6 & \textbf{16.6}$\pm$0.1 & \textbf{12.4}$\pm$0.4 & 40.6$\pm$0.3 & 19.9$\pm$1.2 & \textbf{11.9}$\pm$0.4 \\ \bottomrule
\end{tabular}
}
\caption{The results of entailment tree generation on EntailmentBank test split.
\dag~indicates results from the published paper.
RLET is based on DeBERTa-large, and all other methods are based on T5-large.
}
\label{tab:tree_generation}
\end{table*}}

{
\begin{table}[t]
\small
\centering
\begin{tabular}{@{}lcccc@{}}
\toprule
Method  & FV & SV & HF & Overall \\ \midrule
GPT-3 w/ CoT & \textbf{100.0} & 23.2 & 72.0 & 4.0 \\
ChatGPT  & \textbf{100.0} & 38.3 & 52.0 & 14.0 \\
Entailer & 72.8 & 48.0 & \textbf{82.0} & 24.0 \\ \midrule
\modelname (Ours) & \textbf{100.0} & \textbf{66.0} & \textbf{82.0} & \textbf{46.0} \\ \bottomrule
\end{tabular}%
\caption{
Manual evaluation results on 50 questions randomly sampled from the test split.
FV/SV/HF denotes fact validity/step validity/hypothesis faithfulness.
GPT-3 and ChatGPT have 175B parameters.
Entailer and \modelname are based on T5-large (770M).
}
\label{tab:manual}
\end{table}}

{
\begin{table}[t]
\small
\centering
\begin{tabular}{@{}llll@{}}
\toprule
Algorithm & \multicolumn{1}{c}{All} & \multicolumn{1}{c}{Easy} & \multicolumn{1}{c}{Chal} \\ \midrule
Greedy      & 59.1 & 62.9 & 50.8 \\
Overgenerate-and-filter  & 62.0 & 66.4 & 52.5 \\
Beam search & 64.2    & 67.4      & 57.1      \\ 
Monte-Carlo planning  & \textbf{67.7} &  \textbf{69.5} &  \textbf{63.8} \\ \bottomrule
\end{tabular}%
\caption{
Ablation results of the planning algorithm on the dev split.
All algorithms use the same action budget.
}
\label{tab:search_alg}
\end{table}}

{
\begin{table}[t]
\small
\centering
\begin{tabular}{@{}ccc|ccc@{}}
\toprule
    & \multicolumn{2}{c|}{State Verifier} & \multicolumn{3}{c}{Answer Accuracy} \\ 
    & \texttt{Valid}          & \texttt{Faithful}         & All       & Easy       & Chal       \\ \midrule
(a) & $V_s$         & $V_s$+$V_h$       & \textbf{67.7} &  \textbf{69.5} &  \textbf{63.8}  \\
(b) & $V_s$         & \usym{2715}       & 52.9 &  55.1 &  48.3  \\
(c) & \usym{2715}   & $V_s$+$V_h$       & 61.0 &  62.5 &  57.6  \\
(d) & $V_s$         & $V_h$             & 62.6 &  65.2 &  56.8  \\
(e) & $V_s$         & $V_s$             & 64.2 &  67.2 &  57.6  \\ \bottomrule
\end{tabular}%
\caption{
Ablation results of the state verifier (equation~(\ref{equ:verifier})) on the EntailmentBankQA dev split.
$V_s$ is the step verifier and $V_h$ is the sentence similarity scorer.
}
\label{tab:verifier}
\end{table}}

\noindent \textbf{Manual Evaluation on EntailmentBankQA.}
To make a more accurate investigation, we manually evaluate the quality of trees for the options \textit{selected by models}.
We evaluate along three dimensions.
\textbf{Fact Validity (FV)}: Are the leaf facts correct in the real world?
\textbf{Step Validity (SV)}: Are the intermediate steps valid entailments? 
A step is considered invalid if its conclusion does not follow from the premises or trivially repeats one of the premises.
\textbf{Hypothesis Faithfulness (HF)}: Can the conclusion support the selected answer (even if the answer is not the correct option)?
\textbf{Overall}: The overall score is 1 if all the facts and steps are valid and the conclusion supports the answer.
We invite three students as experts, and the inter-annotation agreement (Cohen's $\kappa$) for FV/SV/HF is 0.726/0.704/0.811.

We also investigate whether FQA can be solved by very large language models, such as GPT-3~\cite{DBLP:conf/nips/BrownMRSKDNSSAA20} and ChatGPT\cite{schulman2022chatgpt}.
For GPT-3, we use Chain-of-Thought prompting (CoT) by treating the entailment tree as the thought. 
For ChatGPT, we describe the task in detail.
Discussion and examples of prompts are in Appendix~\ref{sec:prompts_gpt}.
GPT-3/ChatGPT achieves an answer accuracy of 86\%/92\%, showing that our prompts could appropriately elicit the model's reasoning ability.\footnote{We use the \texttt{text-davinci-003} model for GPT-3 and \texttt{Dec 15 Version} for ChatGPT.}

Table~\ref{tab:manual} shows the results.
We can make the following observations.
(1) While achieving high accuracy rates, LLMs struggle to produce valid and faithful steps, e.g., only 38.3\% of ChatGPT's steps are valid.
(2) Entailer achieves a high HF score but a low FV score since it is not grounded in a fact corpus and may hallucinate possibly incorrect facts to support its answer.
(3) \modelname can answer questions faithfully (HF=82.0\%) based on valid reasoning steps (SV=66.0\%) and human-confirmed facts, achieving the highest overall score.

{
\begin{figure}[t]
\centering
\includegraphics[width=0.8\columnwidth]{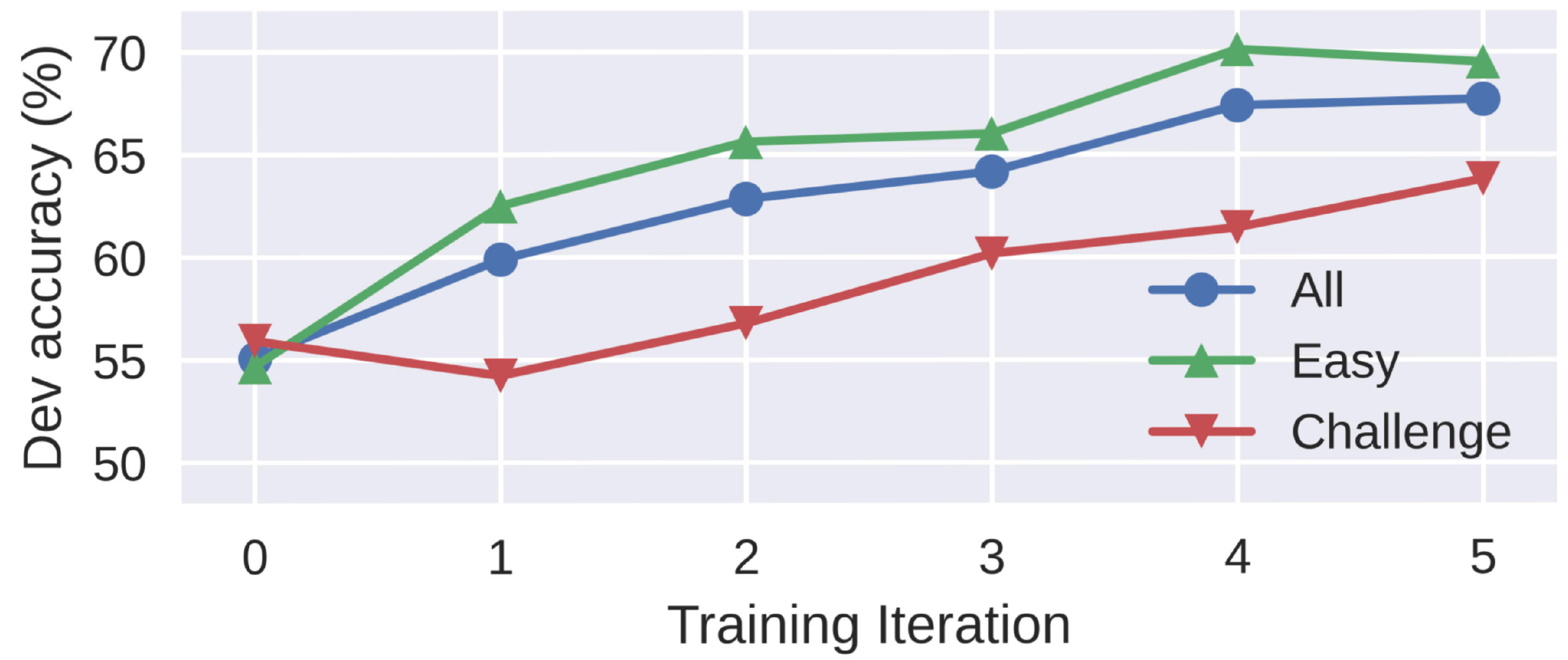}
\caption{Ablation results of the iterative training.
}
\label{fig:iter}
\end{figure}}

\subsection{Ablation Study}

\noindent \textbf{Planning Algorithm.}
To investigate the effectiveness of MCP, we design three comparison algorithms. 
\textbf{Greedy} algorithm selects the action with the maximum prior likelihood score for each state.
\textbf{Overgenerate-and-filter}~\cite{entailer2022} generates $K=5$ candidate actions, executes the actions to get the next states, and selects the one with the highest state score.
\textbf{Beam search}~\cite{DBLP:journals/corr/abs-2208-14271} maintains the best $B$ states.
At each time step, it generates $K$ actions for each state, executes the actions, and keeps the top-$B$ best states for further reasoning.\footnote{We set the beam size $B$ to 3, which achieves the best result. Appendix~\ref{sec:beam_para} includes more discussion of the parameters.}
We set an action budget of 30 for all algorithms.
The algorithms stop when the action budget is used up.
Results in Table~\ref{tab:search_alg} show that MCP improves performance over the comparison algorithms, especially for the challenge questions (from 57.1\% to 63.8\%).\footnote{We also discuss the performance broken down by the length of gold steps in Appendix~\ref{sec:dev_acc_by_depth}.}
MCP could do the look-ahead search efficiently and select actions foresightedly, so as to perform better.

\noindent \textbf{State Verifier.}
As shown in Table~\ref{tab:verifier}, verifying both the step validity and hypothesis faithfulness is necessary (comparing (a), (b), and (c)).
We also find that the ensemble of $V_s$ and $V_h$ yields a more accurate faithfulness score and thus improves performance (comparing (a), (d), and (e)).

\noindent \textbf{Iterative training.}
Figure~\ref{fig:iter} shows that the performance improves with the increasing number of verifier-guided training iterations.
We randomly sample 100 reasoning trajectories selected by the verifier and find that 61\% of them obtain the same trees as the gold ones, 30\% obtain different but still valid trees, and 9\% obtain invalid trees, demonstrating the effectiveness of iterative training.

\section{Conclusion}
We propose \modelname for faithful question answering.
It formulates the task as a discrete decision-making problem and tackles it with a modular reasoning environment and a controller.
\modelname uses the Monte-Carlo planning to make more informed and foresighted decisions.
Experiments show that \modelname can answer questions based on valid and faithful steps and achieve \rw{advanced} performance.

\section{Acknowledgements}
\rw{We appreciate the anonymous reviewers for their insightful comments.
We would like to thank Qihan Liu for the helpful discussions.}
\rw{This work is supported by National Key Research and Development Program (2020AAA0107800) and the Guoqiang Institute of Tsinghua University (2020GQG0005).
}

\section*{Limitations}

Despite outperforming previous best methods, our method still has several limitations and substantial room for future improvement.

First, the variety of modules is limited in the current reasoning environment.
It would be interesting to introduce a wider variety of modules (e.g., a numerical calculator) to make our method more general.

Second, our method currently retrieves facts from a fixed corpus. 
While this is efficient for the specific domain, it may not be sufficient for questions not covered by the fact corpus.
It would be more powerful if we retrieve up-to-date information using a modern search engine as our retriever.

Third, in our experiments, we try our best to select the appropriate prompts to motivate GPT-3 and ChatGPT to generate reasoning steps and answers. 
With our prompts, GPT-3 and ChatGPT can achieve high answer accuracy.
But it is hard to guarantee that our prompts are the best ones to elicit the model's capabilities completely.

Finally, although scaling up the size of the language models may lead to emergent abilities~\cite{DBLP:journals/corr/abs-2206-07682}, in this paper, we do not experiment with larger language models (e.g., T5-11B) due to the computational constraints.

To the best of our knowledge, our work is foundational research, and we do not find obvious risks related to malicious harmful effects, environmental impact, fairness considerations, or privacy considerations.

\bibliography{anthology,custom}
\bibliographystyle{acl_natbib}

\newpage
\clearpage

\appendix
\section{Implementation Details}
\label{sec:appendix_implementation_details}

\noindent $\bullet$ \textbf{Retriever.}
Given a positive pair $(q, k^{+})$, we train the retriever to maximize the cosine similarity between the query $q$ and the positive fact $k^{+}$ in the embedding space.
For an entailment tree in the EntailmentBank training split, we use the hypothesis as $q$ and the leaf facts as $k^{+}$.
The contrastive loss function we used is 
\begin{equation}
    \small
    \mathcal{L} = - \log \frac{\exp (\cos{(q,k^{+})}  / \tau )}{\sum_{i=0}^{bs} \exp (\cos{(q,k_i)} / \tau )},
\end{equation}
where $bs$ is the batch size, $k_i$ is the fact for the $i$ pair in this batch, and $\tau$ is the temperature factor.
We set $\tau$ to 0.02.
The retriever is trained with a learning rate of 1e-5 and a batch size of 40 for 10k steps.
For each retrieval, we return the top 25 facts.
\rw{To update the premise set $X$, we keep all the intermediate conclusions $\mathrm{int}_*$ and replace all other sentences $\mathrm{sent}_*$ with the newly retrieved sentences.
The query sentence is also added to $X$ if it is not in $X$.
}
If a query is retrieved consecutively, we scroll down the retrieval results and return the next 25 facts.

\noindent $\bullet$ \textbf{Entailment Module.}
We follow~\citet{DBLP:conf/naacl/HongZYZ22} to implement the entailment module in a prefixed manner.
All types of modules are implemented with a single T5-large model~\cite{DBLP:journals/jmlr/RaffelSRLNMZLL20}.
A type-specific prefix (e.g., "deductive substitution:") is added to specify which type of reasoning the model should perform.
The model is trained on the entailment steps of the EntailmentBank training split.
The reason type labels of steps are provided by~\citet{DBLP:conf/naacl/HongZYZ22}.
Following~\citet{yang2022nlproofs}, we take the hypothesis as additional input to encourage the module to generate a more relevant conclusion.
The module is trained with a learning rate of 3e-5 and a batch size of 20 following~\citet{DBLP:conf/naacl/HongZYZ22}.

\noindent $\bullet$ \textbf{Controller.}
For each iteration of the verifier-guided iterative training, we train the controller with a learning rate of 1e-5 and a batch size of 20 for ten epochs.
We use the Adafactor optimizer~\cite{DBLP:conf/icml/ShazeerS18}. 
The input sequence of the controller is the concatenation of the hypothesis $H$, partial proof tree ${T}_p$, and context fact set $X$.
The question and option are also included in the input.
An example input is as follows:

\vspace{+2mm}
\noindent
\begin{minipage}{\columnwidth}
\small
{\$question\$ Which will most likely cause a decrease in predator populations? \$option\$ a decrease in prey populations. \$hypothesis\$ a decrease of prey populations will decrease predator populations. \$proof\$ sent1 \& sent2 -> int1 \$context\$ int1: a decrease of food has a negative impact on organisms. sent3: if an organism eats something then that something is a source of frood to that organism. sent4: negative impacts on organisms / species will decrease the population of the organisms / the species. $\dots$ sent25: an adaptation has a positive impact on a living thing 's survival.}
\end{minipage}
\vspace{+1mm}

The action likelihood score is calculated by $\exp(\frac{1}{N} \sum_{i} \text{logit}(x_i|x_{k<i}))$, where $x_i$ is the $i$-th token of the action sequence and $N$ is the number of tokens in the action sequence.

\noindent $\bullet$ \textbf{Step Verifier $V_s$.}
The human-labeled data provided by~\citet{entailer2022} contains 1,827 valid steps and 1,564 invalid steps.
We also use the 4,175 valid steps extracted from the gold trees in the EntailmentBank training split.
For each valid step, we perturb it to make an invalid step by substituting one premise with a random distractor from the corpus.
The step data is randomly divided into training and development splits in the ratio of 9:1.
We train the step verifier with a learning rate of 1e-5 and a batch size of 16 for ten epochs.
An example input of step verifier is as follows:

\vspace{+2mm}
\noindent
\begin{minipage}{\columnwidth}
\small
premises: a mushroom is a kind of fungus. in the food chain process fungi have the role of decomposer. 
conclusion: in the food chain process mushrooms have the role of decomposer.
\end{minipage}
\vspace{+1mm}

In addition, for the sentence similarity scorer $V_h$ in the state verifier, we use \texttt{BLEURT-Large-512} following~\citet{DBLP:conf/emnlp/DalviJTXSPC21}.

\noindent $\bullet$ \textbf{Entailment Tree Construction.}
During reasoning, if adding a step makes the current tree not acyclic, then we consider this step invalid.
For the final state, if the steps form a forest (containing multiple entailment trees), then we pick the tree with the highest hypothesis faithfulness score and discard the other trees.

\noindent $\bullet$ \textbf{Total Number of Learnable Parameters.}
In our \modelname, the numbers of learnable parameters of the controller, the retriever, the entailment module, and the step verifier are 770M, 109M, 770M, and 435M, respectively.
In total, the number of learnable parameters of \modelname is about 2,084M.

\noindent $\bullet$ \textbf{Experiment Environments.}
We deploy all models on a server with four RTX 3090 GPUs.
Each RTX 3090 GPU has 24G of memory.
Our code mainly depends on python 3.8.0 and PyTorch 1.10.0.
We use the pre-trained language models from \texttt{HuggingFace Transformers}\footnote{\url{https://github.com/huggingface/transformers}}.

\noindent $\bullet$ \textbf{Running time.}
In our experiment environment, the average running time for each option is 30.77$\pm$5.93 seconds.
This running time is acceptable.
While not directly comparable, the running time for each option is about 90 seconds for NELLIE~\cite{DBLP:journals/corr/abs-2209-07662} and 20$\sim$60 seconds for Entailer-11B~\cite{entailer2022}, as reported in their papers.          
\section{\modelname Illustrations and Case Study}
\label{sec:Illustrations}
\begin{figure*}[t!]
\centering
\includegraphics[width=\textwidth]{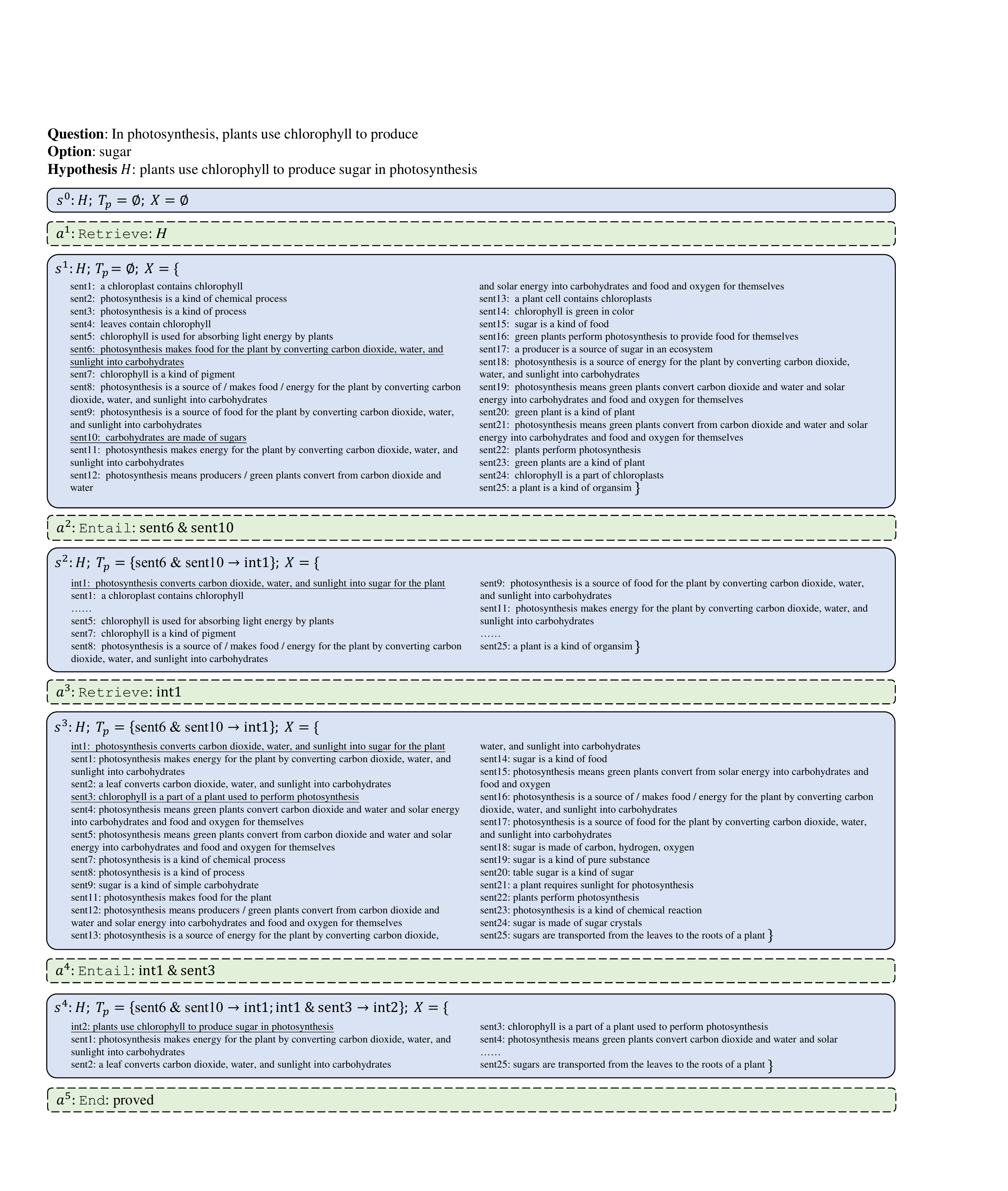}
\caption{
An illustration of the reasoning process of \modelname.
For a question and the option that we want to prove, we first convert the question+option to a hypothesis $H$.
We start from the initial reasoning state where the partial tree $T_p=\emptyset$ and the candidate premises $X=\emptyset$.
In each interaction, we execute an action and update the reasoning state.
For the action \Action{Retrieve}, we sent the query to the retriever and update $X$ with the retrieval results.
For the action \Action{Entail}, we sent the selected premises to the entailment modules.
The novel conclusion is added to $X$ and the novel step is added to $T_p$.
For the action \Action{End}, we end the reasoning process.
}
\label{fig:trace_example}
\end{figure*}

\begin{figure*}[t!]
\centering
\includegraphics[width=\textwidth]{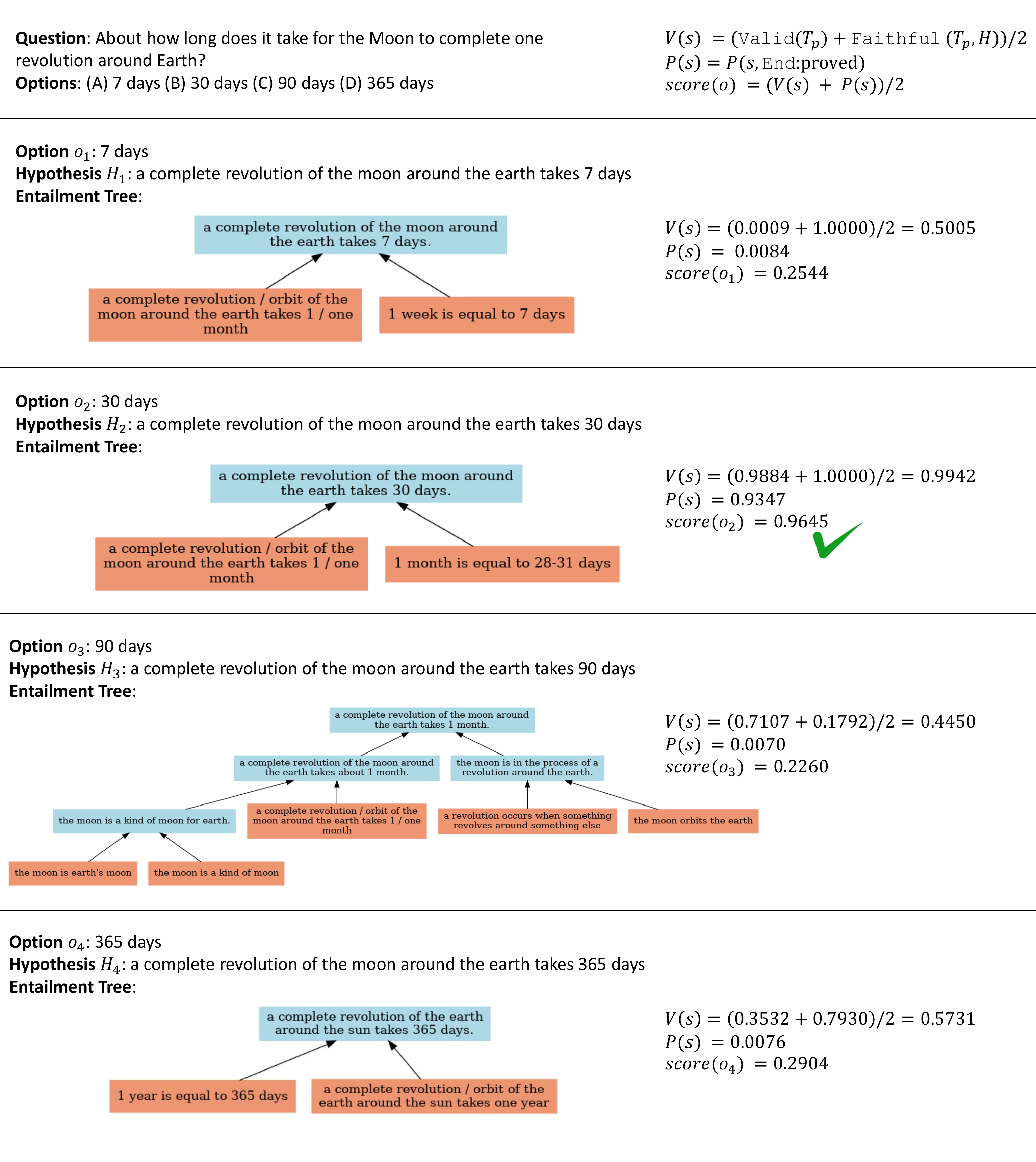}
\caption{
An example of the option selection.
For each option, we try to generate an entailment tree to prove the option.
We select the option based on the state verifier score $V(s)$ and the controller score $P(s)$.
}
\label{fig:tree_for_choices_example}
\end{figure*}

\begin{figure*}[t!]
\centering
\includegraphics[width=\textwidth]{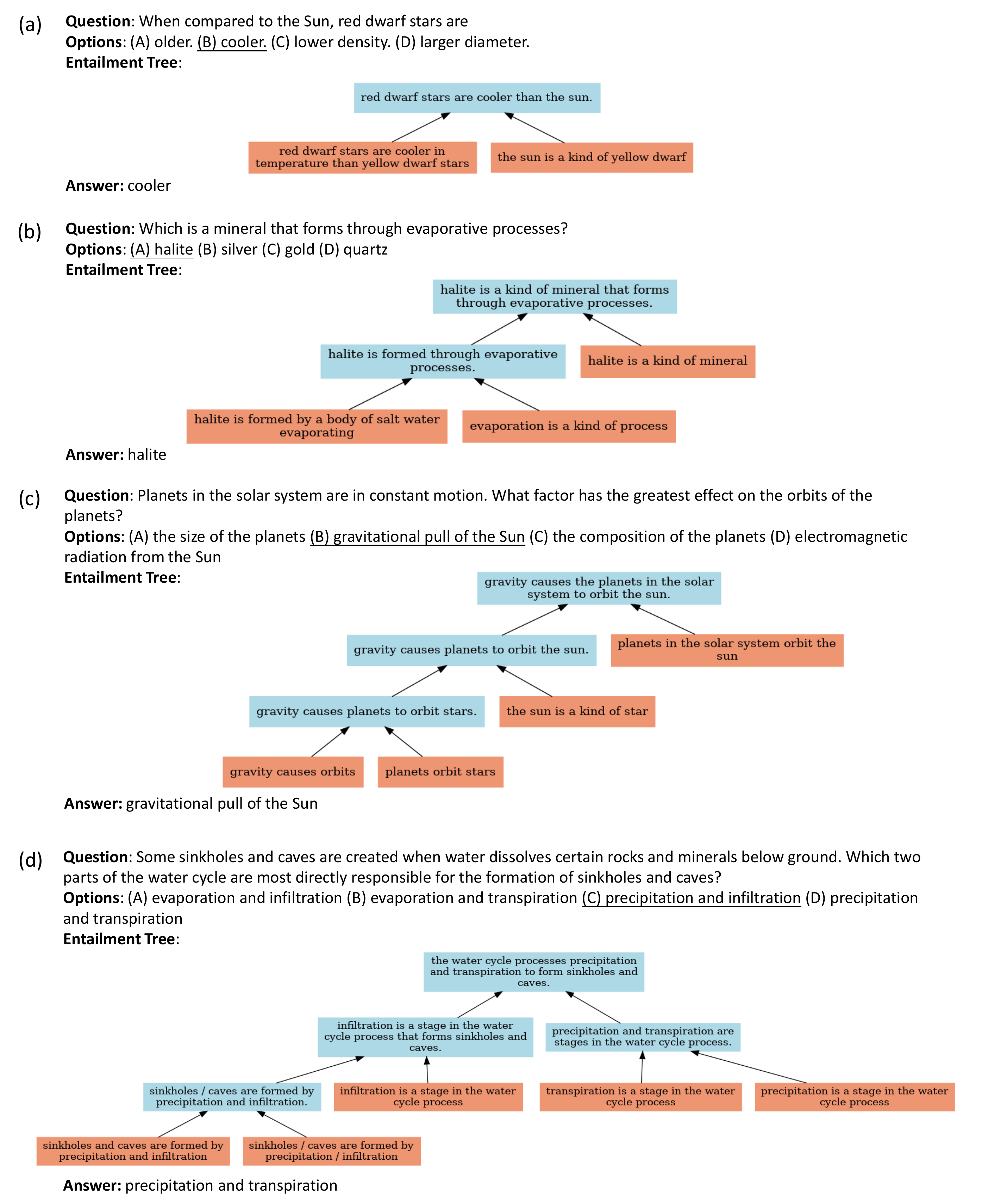}
\caption{Reasoning steps and answers generated by \modelname.
The correct options are indicated by underlining.
In cases (a), (b), and (c), \modelname can generate valid entailment trees and eventually select the correct options.
In case (d), \modelname selects the wrong option.
The causes of the selection errors are traceable.
Some steps in the entailment tree are invalid, leading to the wrong reasoning and the wrong option.
}
\label{fig:cases}
\end{figure*}

Given a question and candidate options, we first try to prove each option by generating an entailment tree (Figure~\ref{fig:trace_example}).
Then, we score each option and select the option with the highest score (Figure~\ref{fig:tree_for_choices_example}).
Figure~\ref{fig:cases} shows some entailment trees and answers generated by our method on the EntailmentBankQA test split.

\section{Additional Experiment Results}

\subsection{Planning Algorithm Hyperparameter Analysis}
\label{sec:mcts_para}
\begin{figure}[t!]
\centering
\includegraphics[width=\columnwidth]{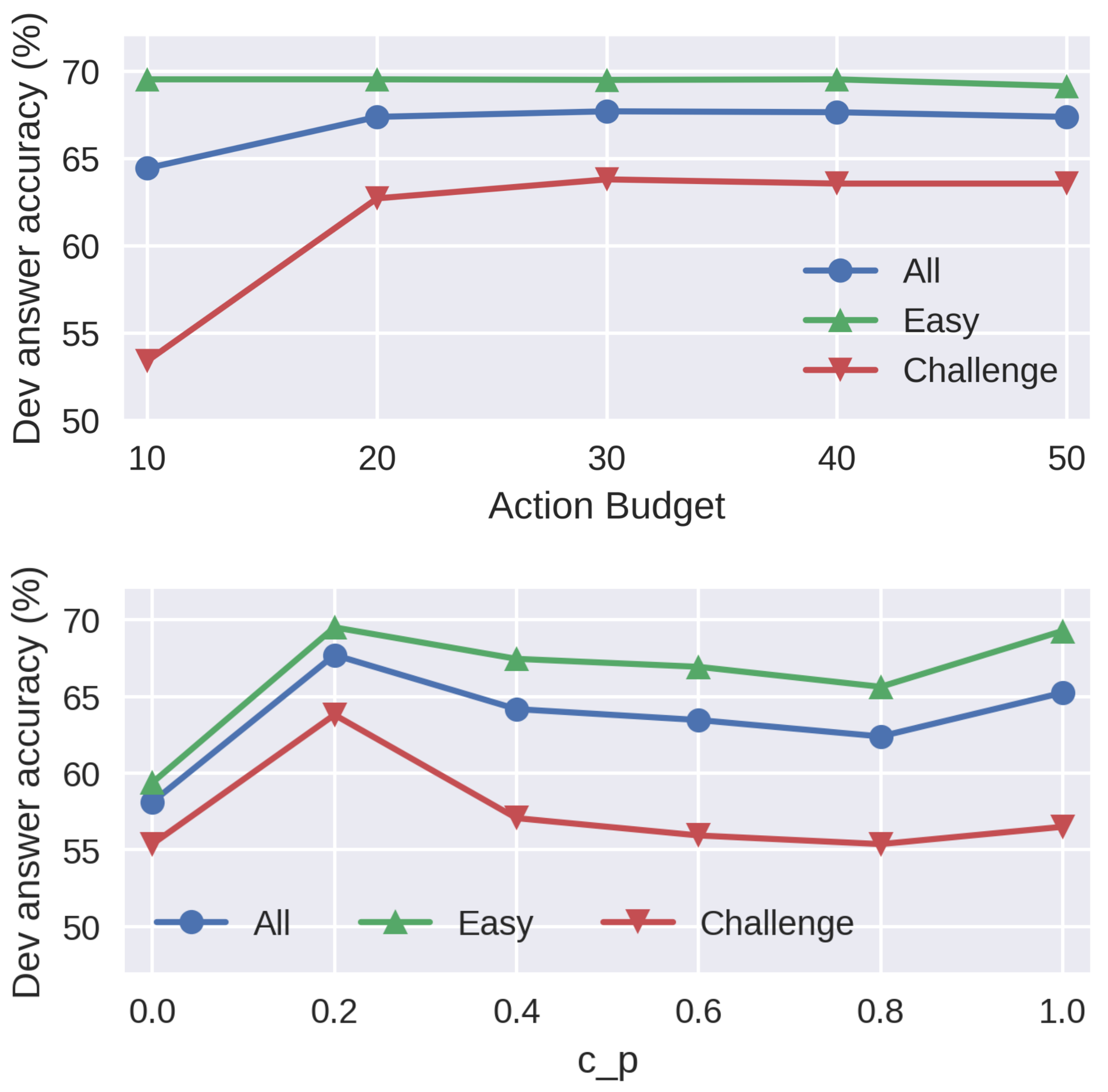}
\caption{
Results of Monte-Carlo planning algorithm with different parameters on EntailmentBankQA dev split.
}
\label{fig:mcts_para}
\end{figure}

The hyperparameters are selected using the dev split, as shown in Figure~\ref{fig:mcts_para}.
We select an action budget of 30 and $c_p=0.2$.
We can also find that, as the action budget increases, the performance remains basically unchanged on easy questions but improves on challenge questions.

\subsection{Beam Search Hyperparameter Analysis}
\label{sec:beam_para}
Table~\ref{tab:beam_para} shows the beam search performances with different beam sizes and action budgets.
In the case of an action budget of 30 (the same as other algorithms), a beam size of 3 achieves the highest answer accuracy.
We also find that simply increasing the action budget brings limited performance gains,  especially for the challenge questions.

\begin{table}[t]
\small
\centering
\begin{tabular}{@{}cc|ccc@{}}
\toprule
Beam  size & Action budget & All & Easy & Chal \\ \midrule
2           & 30     & 63.6    & 66.4      & 57.6          \\
3           & 30     & 64.2    & 67.4      & 57.1           \\
5           & 30     & 62.8    & 65.2      & 57.6           \\ \midrule
3           & 50     & 65.1    & 68.2      & 58.2           \\
3           & 100    & 65.0    & 68.2      & 58.1           \\ \bottomrule
\end{tabular}%
\caption{
Results of beam searches with different parameters on EntailmentBankQA dev split.
}
\label{tab:beam_para}
\end{table}

\subsection{Performances Breakdown}
\label{sec:dev_acc_by_depth}
\begin{figure}[t]
\centering
\includegraphics[width=\columnwidth]{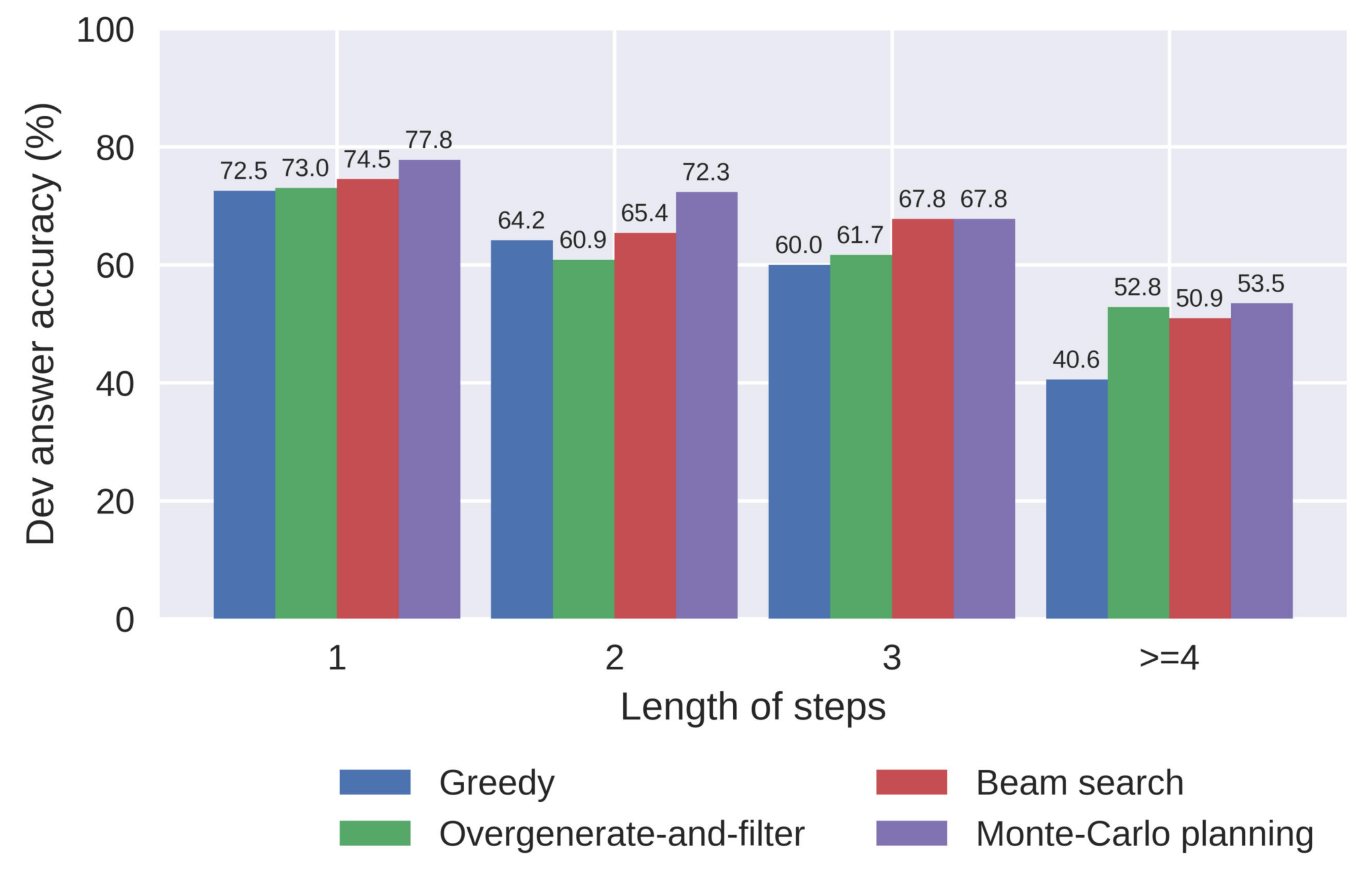}
\caption{
Dev results of different planning algorithms broken down by the length of the steps in the gold trees.
}
\label{fig:dev_acc_by_depth}
\end{figure}

Figure~\ref{fig:dev_acc_by_depth} shows the performances of different planning algorithms on the EntailmentBankQA development split.
We break down the results by the length of steps in the gold trees of the correct options.
The results show that the questions become increasingly difficult as the length of steps increases.
The Monte-Carlo planning algorithm achieves better results than other algorithms at all lengths.

\subsection{Error Analysis}
\label{sec:error_analysis}
We randomly sample 50 questions that \modelname answers incorrectly.
We find the following three types of errors.

\textbf{(1) Declarativization Error (32\%).}
Our generation model that converts the question+option to the declarative hypothesis is not free from conversion errors.
The conversion result might lack some key information and thus not match the original question+option.
For example, while the question+option is \textit{The most likely reason sound travels faster in saltwater than in freshwater is that saltwater}+\textit{is more elastic}, the generated hypothesis is \textit{saltwater is likely to travel faster than freshwater}.
Given the mistaken hypothesis, it is difficult for the controller to reason correctly.

\textbf{(2) Invalid Reasoning Step (40\%).}
One of the reasoning steps is invalid, leading to the incorrect conclusion.
A step is invalid if its conclusion does not follow from the premises, e.g., the conclusion is in conflict with or irrelevant to the premises.
An example of an invalid step is \textit{planting trees has a positive impact on an environment \& negative impact is the opposite of positive impact -> planting a tree has a negative impact on the environment}.

\textbf{(3) Unsupported Option (28\%).}
The final conclusion of the reasoning steps is not relevant to the selected option or does not contain enough information to support the selected option, even if the reasoning steps are valid.
For example, the hypothesis is \textit{granite can be used to study the history of organisms}, while the final conclusion of the reasoning steps is \textit{granite can be used to study the history of rocks on earth}.

\section{Prompts for GPT-3 and ChatGPT}
\label{sec:prompts_gpt}
\begin{figure*}[t!]
\centering
\includegraphics[width=0.95\textwidth]{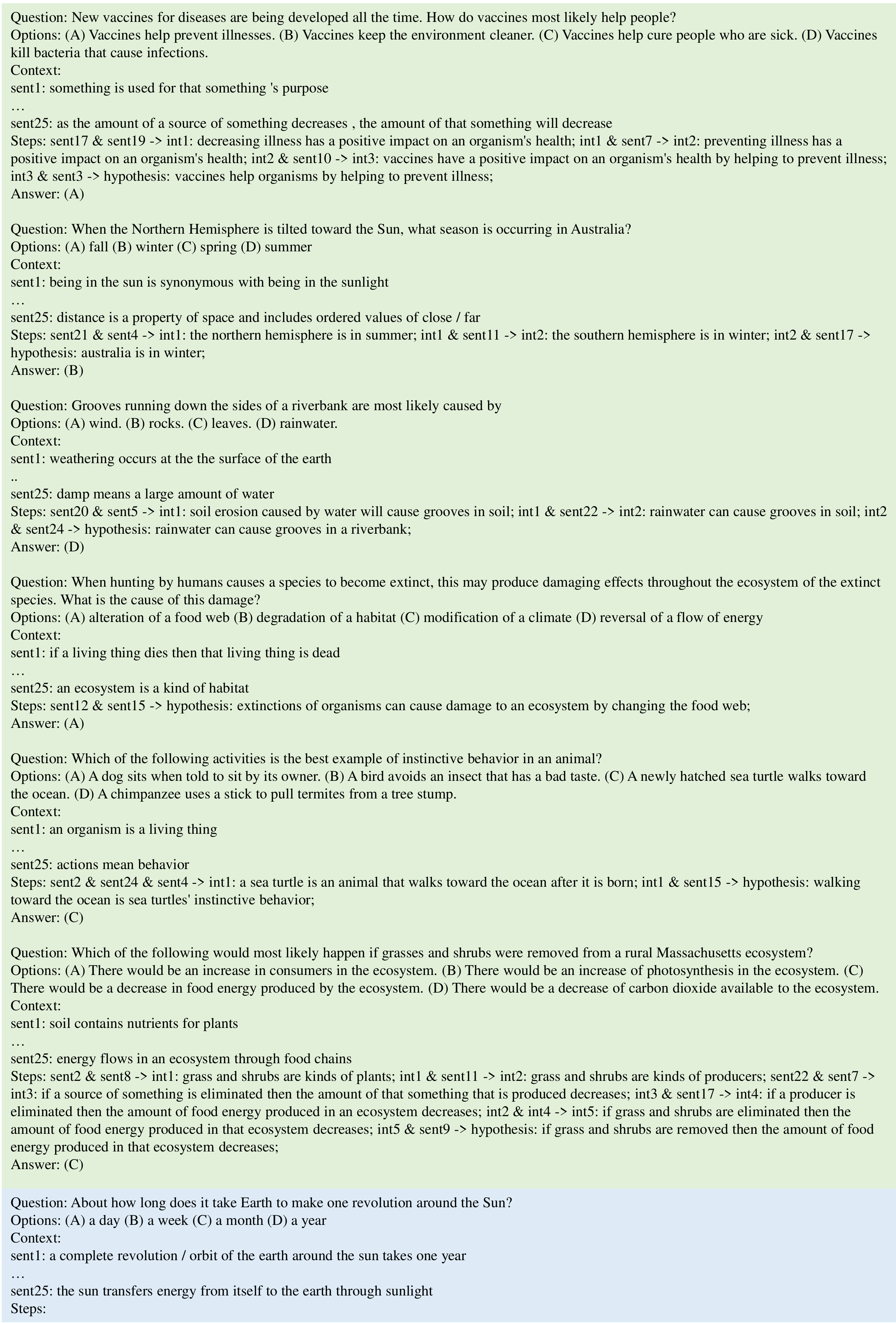}
\caption{
A Chain-of-Thought prompt for GPT-3. 
The prompt consists of six in-context examples (with entailment trees and correct answers) and the test example (without the entailment tree and correct answer).
For simplicity, we show only two facts for each example.
}
\label{fig:gpt_prompt}
\end{figure*}

Figure~\ref{fig:gpt_prompt} shows an example prompt for GPT-3.
We randomly sample six examples (to fit the maximum input of GPT-3) from the training split of EntailmentBank (Task 2).
We use the Chain-of-Thought prompting~\cite{DBLP:journals/corr/abs-2201-11903} by using the entailment tree as the intermediate thought.
For the test example, we retrieve 25 facts from the corpus using the question as the query.
We use the same retriever as our method.

\begin{figure*}[t!]
\centering
\includegraphics[width=0.9\textwidth]{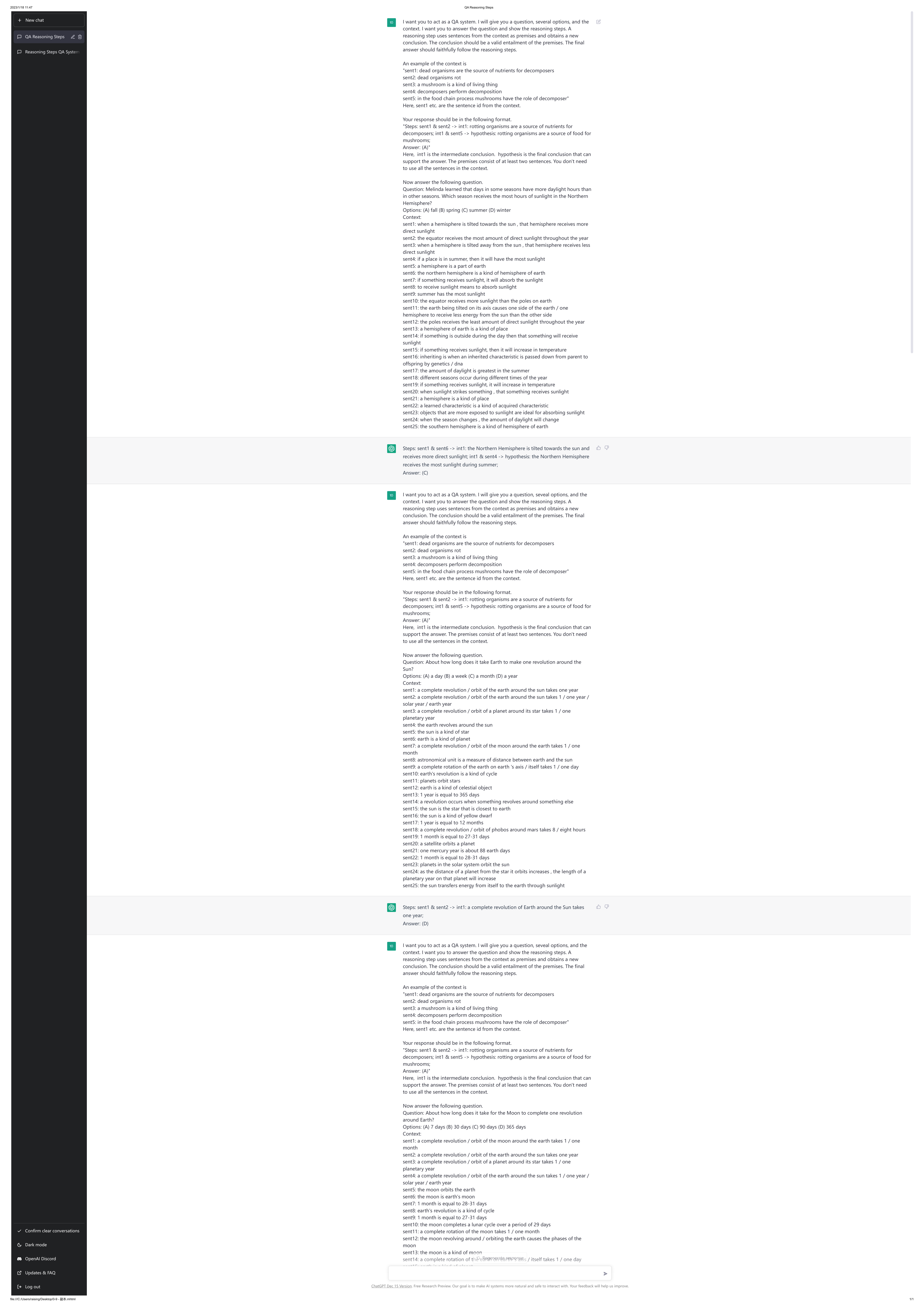}
\caption{
A dialogue example with ChatGPT.
We introduce the task in detail.
ChatGPT can respond in the desired form.
}
\label{fig:chatgpt_prompt}
\end{figure*}

Figure~\ref{fig:chatgpt_prompt} shows a dialogue with ChatGPT.
We introduce the definition of the task in detail and guide the model to respond in the desired form.
We manually try several kinds of prompts and chose the best one.
For example, we try to include several in-context examples in the prompt as we do for GPT-3, but it doesn't seem to yield better results.

\section{Automatic Evaluation Metrics on EntailmentBank}
\label{sec:auto_tree_metric}
For the entailment tree generation task, we evaluate on EntailmentBank using their automatic evaluation metrics~\cite{DBLP:conf/emnlp/DalviJTXSPC21}.
Denote the predicted entailment tree as $\hat{{T}}$ and the gold one as ${{T}}$.
First, a tree alignment algorithm is performed to align nodes of $\hat{{T}}$ to nodes of ${{T}}$.
Leaf nodes are aligned based on their texts.
Each non-leaf node of $\hat{{T}}$ is aligned to the first non-leaf node of ${{T}}$ which has the largest Jaccard similarity of the leaf node.
Then, we evaluate $\hat{{T}}$ with the following metrics:

\noindent $\bullet$ \textbf{Leaves}.
The leaf nodes of $\hat{{T}}$ and ${{T}}$ are compared to compute F1 score.
The AllCorrect score is 1 if F1 is 1, 0 otherwise.

\noindent $\bullet$ \textbf{Step Structures}.
To evaluate whether the steps of $\hat{{T}}$ are structurally correct, the steps of $\hat{{T}}$ and ${{T}}$ are compared to compute the F1 score.
A predicted step (corresponding to a non-leaf node of $\hat{{T}}$) is structurally correct if its children nodes perfectly match the gold ones.
AllCorrect=1 if F1=1, 0 otherwise.

\noindent $\bullet$ \textbf{Intermediates}.
To evaluate the predicted intermediate conclusions, the intermediate F1 score is computed by comparing the aligned intermediate conclusions.
A predicted intermediate conclusion is considered correct if the \texttt{BLEURT-Large-512} score of the aligned conclusion pair is larger than 0.28.
AllCorrect=1 if F1=1, 0 otherwise.

\noindent $\bullet$ \textbf{Overall AllCorrect}.
Based on the above scores, the Overall AllCorrect score is 1 if and only if the AllCorrect scores of Leaves, Step Structures, and Intermediates are all 1.

Please refer to the EntailmentBank paper for more details.

\newpage

\end{document}